\documentclass[letterpaper, 10 pt, conference]{ieeeconf}  

\IEEEoverridecommandlockouts                              

\usepackage{graphics} 
\usepackage{epsfig} 
\usepackage{mathptmx} 
\usepackage{times} 
\usepackage{amsmath} 
\usepackage{amssymb}  
\usepackage{subfigure}
\usepackage{makecell}
\usepackage{graphics}
\usepackage{cite}
\usepackage{multirow}
\usepackage{url}
\usepackage{verbatim} 
\usepackage{hyperref} 
\usepackage{stfloats}
\usepackage[table]{xcolor}
\usepackage{comment}
\usepackage{tabularx}

\title{\LARGE \bf
LiePoseNet: Heterogeneous Loss Function Based on Lie Group for Significant Speed-up of PoseNet Training Process 
}

\author{Mikhail Kurenkov, Ivan Kalinov and Dzmitry Tsetserukou
\thanks{All authors are with Skolkovo Institute of Science and Technology, Moscow 143026, Russia 
        {\tt\small\ \{mikhail.kurenkov, d.tsetserukou\} @skoltech.ru, ivan.kalinov@skolkovotech.ru}}%
}

\begin{document}
\maketitle
\thispagestyle{empty}
\pagestyle{empty}

\begin{abstract}
Visual localization is an essential modern technology for robotics and computer vision. Popular approaches for solving this task are image-based methods. Nowadays, these methods have low accuracy and a long training time. The reasons are the lack of rigid-body and projective geometry awareness, landmark symmetry, and homogeneous error assumption. We propose a heterogeneous loss function based on concentrated Gaussian distribution with the Lie group to overcome these difficulties. Following our experiment, the proposed method allows us to speed up the training process significantly (from 300 to 10 epochs) with acceptable error values.
\end{abstract}

\section{Introduction}
\subsection{Motivation}

Visual localization is an essential part of robotic frameworks. It allows a robot to estimate its position and confidently operate in different environments. The current gold standard for localization are LIDAR-based methods. They have high accuracy, but their main limitation is the LIDAR sensor itself, that is expensive and outputs sparse point clouds. Visual localization is a promising alternative due to its low-cost and dense sensors. Cameras have fewer blind zones, rich information output, and a wide field of view. Thus, it can reduce the hardware setup cost and lead to the large-scale proliferation of robots in all areas of our lives.

Modern robotics significantly demands large-scale, long-term robust visual localization methods. For large-scale visual localization, the model size is an essential requirement. However, for long-term localization, it is essential to have methods with fast learning time because of quick adaption to new environment changes (season and lighting changes, day/night, dynamic objects)\footnote{\url{https://www.visuallocalization.net/}}. Moreover, these methods have to be robust to real scenes with complex structures, such as ambiguous objects and symmetric landmarks. These problems lead to low accuracy of the visual localization methods and limit their usage for robots in real environment, for example, for delivery \cite{karpyshev2022mucaslam, protasov2021cnn}, warehouse inventory \cite{kalinov2019high, kalinov2020warevision, kalinov2021warevr, kalinov2021impedance} or even disinfection \cite{perminov2021ultrabot, mikhailovskiy2021ultrabot}.

\subsection{Problem Statement}
\label{Problem}

Chen et al. \cite{chen2020survey} presented a taxonomy of visual localization methods, according to which the most accurate are: image-based (2D-2D) and structure-based (2D-3D) methods. Structure-based methods, such as \cite{brachmann2016, sattler2017, brachmann2018learning, schonberger2018}, propose the use of a 3D model of the environment. They firstly match 2D pixels to this 3D model during operation and then estimate the camera position using PnP algorithm. Methods \cite{donoser2014, kendall2015, kendall2016, clark2017, Kendall2017, walch2017, melekhov2017, melekhov2017relative, acharya2019} have an explicit or implicit geotagged image database, and during operation, they match images to this database and then estimate the camera position by interpolation. Structure-based methods are more accurate than image-based, but image-based approaches are faster during operation.

One of the most popular approach for image-based localization is an absolute pose regression method \cite{kendall2016, kendall2015, Kendall2017}. It performs pose estimation by only a single pass of artificial neural networks. These algorithms do not require any hand-crafted features and are also trained in an end-to-end manner. Nonetheless, their main disadvantages are low accuracy, long training time, big model size, and the necessity of ground truth positions for training.

One of the main reasons for the low accuracy of pose regression methods is the need for more awareness about rigid-body and projective geometry. Structure-based methods can use knowledge about projective geometry, but absolute pose regression methods need to learn from scratch. Due to this, and the insufficient training data, pose regression methods are overtrained on the training dataset \cite{Sattler2019}. It leads to poor generalization ability. Loss function can give some knowledge about geometry, but reasonable priors must be established. Such loss function would allow faster learning due to an improved understanding of the environment geometry.  

\begin{figure}[!t]

\begin{center}
        \subfigure[Cylindrical symmetric object.]{
        \resizebox*{4.1cm}{!}{\includegraphics{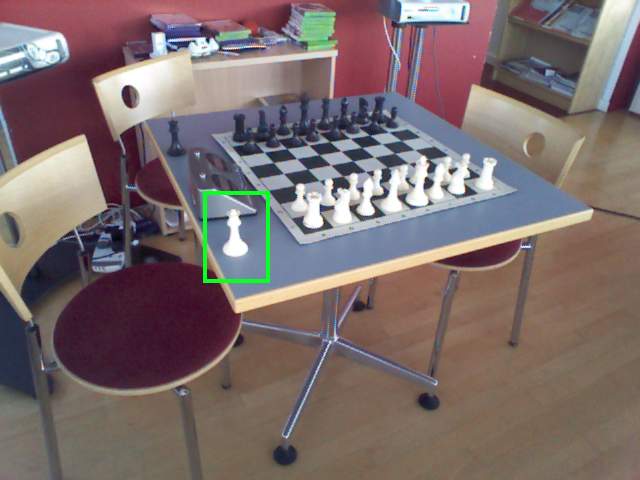}}}
        \vspace{-0.5em}
        \subfigure[Linear symmetric object.]{
        \resizebox*{4.1cm}{!}{\includegraphics{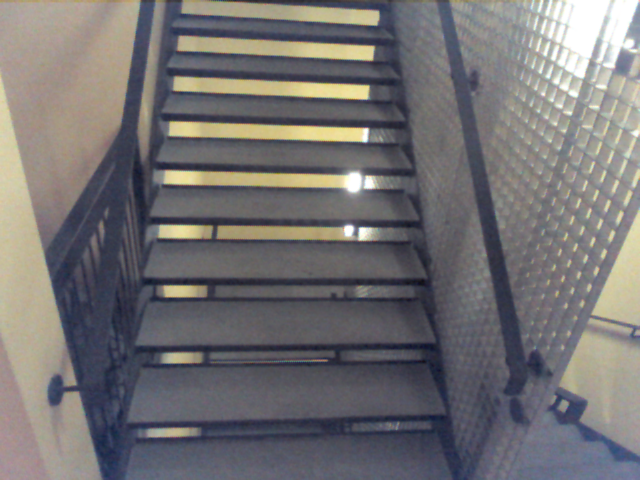}}} 
        \caption{\label{fig:validation_results} Examples from 7-Scenes dataset with different types of symmetry.}
        \label{fig1}
\end{center}
\vspace{-5ex}
\end{figure}
Another problem with the absolute pose regression method is a scene ambiguity due to symmetry of landmarks. This symmetry can be linear, cylindrical, or spherical. Examples of a symmetric object are shown in Fig. \ref{fig1}(a) from 7-Scenes dataset\footnote{\url{https://www.microsoft.com/en-us/research/project/rgb-d-dataset-7-scenes}}. It is possible to see a good landmark figure, but it has cylindrical symmetry. Another example of symmetry is the linearly symmetrical object shown in Fig. \ref{fig1}(b). Not all images allow estimating the camera position with equal accuracy. However, modern approaches proposed homogeneous errors for all examples in the training dataset. Therefore, every method could be overtrained if bad image examples are provided for it. As a result, this greatly complicates the learning process and significantly increases the training time.

\subsection{Related Work}
Visual localization is the task of determining the position of a camera from known data. Modern visual localization methods use geometric primitives \cite{donoser2014, meng2017, sattler2017, schonberger2018} and consist of two steps. The first step is to find a match between geometric primitives in 2D images and 3D maps. The second step is calculation of the final camera position. Geometric primitives can be either key points, as in ORB-SLAM \cite{donoser2014}, or image areas \cite{meng2017}. The comparison is carried out for key points using descriptors. Calculation of the camera final position is a classic PnP problem, in which the RANSAC method is often used to stabilize the solution. The current research line is building a complete neural network that integrates all of these steps, it is called the end-to-end approach.

The end-to-end approach has been increasingly used recently. This approach determines the camera coordinates based on the one image at the one pass of the neural network. The difference between this task and visual terrain recognition is that the terrain recognition finds similar places where the robot was. The visual localization task finds the exact coordinate of the robot, frequently from an already existing three-dimensional map, which is explicitly set in the form of a point cloud or implicitly in the form of a neural network. Visual localization methods are described in more detail in the overview \cite{Piasco2018}.  

The task of finding camera coordinates in three-dimensional space using an end-to-end neural network is called absolute pose regression (APR). The neural network is trained to find the coordinates of the camera on a specific scene. In contrast to this problem, the problem of relative camera pose regression (RPR) is highlighted. The camera coordinates are predicted relative to predefined images, not relative to the global coordinate system. This task is general, since the neural network learns not for a specific scene but for arbitrary scenes.  

Kendall et al. introduced the first fundamental work in this area \cite{kendall2015}. The presented neural network architecture for determining the absolute coordinates of the PoseNet camera consists of three layers. At the input of the PoseNet network, an image is supplied, and at the output, the algorithm gives the camera coordinates. The first layer defines local features. This layer uses fully connected parts of the pre-trained architecture, for example, ResNet \cite{he2016}. A non-linear aggregator is considered on the next layer, it converts the image into a high-dimensional vector. Furthermore, the last layer considers the linear projection from the high-dimensional vector to the camera coordinates. In the following work, \cite{kendall2016}, the authors applied Bayesian inference to the neural network for computing the uncertainty of the final PoseNet coordinate. 

Brahmbhatt et al. proposed MapNet algorithm that used the geometric constraints to improve the accuracy of localization by camera \cite{Brahmbhatt2018}. They used labeled and unlabeled data for training and visual odometry as a one of the data sources. They entered the visual odometry into the loss function as the difference between the measured offset and the predicted result. Thus, MapNet works similarly to the classic SLAM algorithm, but the structure of its neural network is similar to PoseNet. 

Walch et al. \cite{walch2017} proposed to improve the PoseNet architecture by adding LSTM layers for reducing the dimension of feature vectors. The LSTM layer was located between the second layer in the described PoseNet architecture, which found a high-dimensional feature vector. Huang et al. improved the PoseNet architecture to work with dynamic objects by using a neural network that ignored the front objects in the image because they were often dynamic \cite{huang2019prior}. Also, the PoseNet accuracy could be improved by using synthetic data \cite{acharya2019}


Brachmann et al. presented the neural networks for partial regression only described in \cite{brachmann2018learning}. The method used two separate neural networks. One network looked for a coordinate for each image area and set the hypotheses using a random selection of four coordinates. The second neural network evaluated the hypotheses, and the final coordinate was obtained as the best hypothesis.

Several papers described uncertainty estimation for deep learning tasks. For example, Yang et al. presented a work \cite{yang2020d3vo} on the deep learning for visual odometry, depth, and uncertainty estimation. Another work \cite{liu2020tlio} proposed a method for inertial odometry, which learned for the prediction of the displacement of IMU measurements and uncertainty based on raw IMU data. These works used the uncertainty estimation for the visual or inertial odometry tasks, not for the absolute pose regression tasks.

It is important to note that all the works listed above aim to improve visual localization accuracy; moreover, the main factor that hinders the development of visual localization methods is the long training time. For example, the training time required to recognize a 2-10  minute video from a dataset could be about one day \cite{Kendall2017}.

\subsection{Contributions}
Our work aims to improve absolute pose regression methods by increasing the accuracy of pose regression algorithms with fixed training time.

We introduce the heterogeneous loss function based on the Lie group to solve problems connected with absolute pose regression. Our heterogeneous loss function models heteroscedastic uncertainty by concentrated Gaussian distributions. This approach has been proven suitable for solving SLAM and visual odometry problems. This loss function has strong prior knowledge about SE(3) geometry. Moreover, it correlates position and rotation errors, which is essential for handling symmetric landmarks. Finally, the proposed loss function is heterogeneous, and this characteristic will improve the learning process on challenging datasets.

To highlight the performance of the proposed approach\footnote{\url{https://github.com/MisterMap/lie-pose-net}} we present experiments on the 7-Scenes dataset and show that this method can achieve the same accuracy as PoseNet, while requiring less epochs for training. Thus, the presented method can significantly speed up the training process.

{\bf Our main contributions in this work are as follows:} 
\begin{itemize}
    \item Heterogeneous loss function based on concentrated Gaussian distributions, which can handle images with symmetric landmarks and allows learning on images with poor quality.
    \item Significant speed-up of the training process with the same accuracy by using the proposed loss function. 
\end{itemize}

\section{Methodology}
\subsection{Concentrated Gaussian Distribution on the Lie Group}

The core idea of our approach is to use SE(3) Lie group theory in the PoseNet loss function. SE(3) group is the special Euclidean group in three dimensions responsible for rigid body transformations. It consists of translation and rotation parts. Lie group or $G$ is a differentiable smooth group. It means that the composition of two elements and the reverse operation is smooth and differentiable. The tangent space to the identity element of the Lie group is a Lie algebra $\mathfrak{g}$. The transformation from the Lie algebra to the Lie group is called exponential mapping, and the reverse transformation is called logarithm mapping. The exponential mapping arises from the integration of velocities into the Lie group. Thus, the Lie group is a natural representation of different geometries from physics, including SE(3) geometry. 

\begin{figure} [!t]
\vspace{1ex}
\begin{center}
\includegraphics[width=8.6 cm]{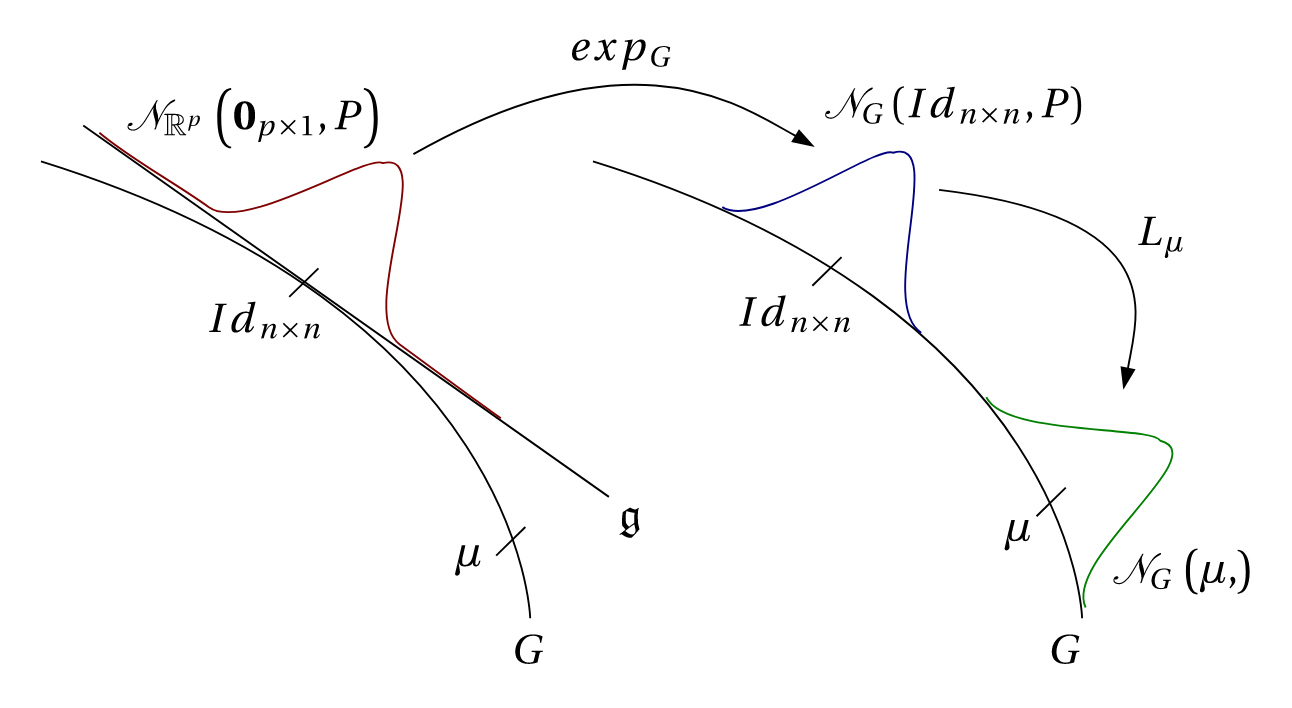}
\vspace{-3em}
\caption{Concentrated Gaussian distribution \cite{Bourmaud2014}.}
\label{fig:concentrated_gaussian_distribution}
\end{center}
\vspace{-2.5em}
\end{figure}

The Lie algebra is applied to represent distribution invariant to the transformation of the Lie group: composition and reverse. One of these distributions is a concentrated Gaussian distribution, shown in Fig. \ref{fig:concentrated_gaussian_distribution}. This distribution is an extension of simple Gaussian distribution for translation and rotation. Concentrated Gaussian distribution is often used for Kalman filter in robotics and for solving the SLAM problem \cite{Bourmaud2014, Lenac2018}. 

In our work, we use concentrated Gaussian distribution to model the predicted uncertainty of position. The logarithm and exponential mapping of the Lie group have analytical formulas and can be differentiated during training. Thus, the proposed loss function is the negative log-likelihood of a concentrated Gaussian distribution.

The sampling from this distribution could be performed by using the following equation:

\begin{equation}
x = \mu \exp_G(\delta),
\end{equation}

where $\delta \sim \mathcal{N}(0, \Sigma), x, \mu \in G, \delta \in \mathfrak{g}$.

The probability of position $x$ can be obtained as follows:
\begin{equation}
    p(x) = \frac{1}{\sqrt{(2 \pi)^N \det|\Sigma|}}e^{-\frac{1}{2}||\log(\mu^{-1} x)||_{\Sigma}^2}
    \label{eq:lie_distribution}
\end{equation}

\subsection{PoseNet}
Original PoseNet \cite{kendall2015} is a convolution neural network (CNN) that predicts camera position from the image obtained by this camera. Let us denote $X$ as the predicted position, $Y$ is the input image, and $\hat{X}$ is the ground truth position. Therefore, our goal is to estimate the maximum likelihood of the given model $p_\theta(X|Y)$.

\begin{figure} [!t]
\vspace{0.5em}
\begin{center}
\includegraphics[width=8.6 cm]{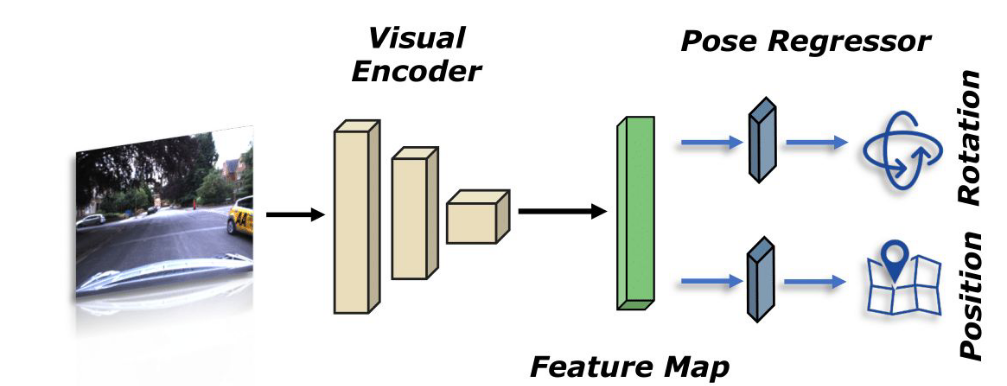}
\vspace{-1em}
\caption{PoseNet structure \cite{chen2020survey}.}
\label{fig3}
\end{center}
\vspace{-2.5em}
\end{figure}

The structure of PoseNet is shown in Fig. \ref{fig3}. The PoseNet consists of the feature extractor, the feature mapping, and the position regressor. Feature extractor is a deep CNN network, for example, ResNet \cite{he2016}, or VGG \cite{russakovsky2015imagenet}. The output of the feature extractor is mapped to the feature dimension, which is then passed to position regression. Usually, two fully connected layers are used: one for the rotation and one for the position. Nevertheless, we used only one fully connected layer with 3 + 4 + 21 output dimensions in our framework. The sequence of dimensions is represented as three dimensions for the position, four dimensions for the rotation, and 21 dimensions for the parametrization of the covariance matrix.

We parametrize position for SE(3) space by the 3D vector translation and the 4D normalized quaternion. After the normalization of the quaternion, we calculate the rotation matrix. This calculation and the calculation of SE(3) logarithm are analytical, and both have the analytical derivative. 

\subsection{Geometric Loss Functions}
Original PoseNet and its following implementations used geometric loss functions \cite{kendall2015, kendall2016, Kendall2017}. Discussing them before introducing our heterogeneous loss function based on the Lie group is essential. The camera position $X$ is parametrized in these functions by translation $x$ and rotation $q$. The quaternion parameterization is used for the rotation.

The optimization of the loss function in original PoseNet \cite{kendall2015} is represented by the equation \ref{eq:posenet_loss}. In this equation, $x$ is the predicted camera coordinate, $\hat{X}$ is the true position, $\hat{q}$ is the true quaternion of the camera orientation, and $q$ is the predicted camera orientation quaternion.

\begin{equation}
\mathcal{L}(Y, \hat{X}) = \left\lVert \hat{x} - x\right\rVert ^ 2 + \beta \left\lVert \hat{q} - \frac{q}{\left\lVert q \right \rVert}\right\rVert ^ 2.
\label{eq:posenet_loss}
\end{equation}

Kendall et al. proposed using learnable parameters $s_x$ and $s_q$ \cite{kendall2015} instead of using the meta parameter $\beta$, which represented the trade-off between the rotation and translation losses. They used a probabilistic approach and treated the loss function as the negative logarithm of the normal distribution. Still, instead of the covariance parameterization of the normal distribution, they chose more stable parametrization $s = \log \sigma^2$, where $\sigma$ is the scaling factor of normal distribution (equation \ref{eq:geom_posenet_loss}).

\begin{equation}
\mathcal{L}(Y, \hat{X}) = e^{-s_x} \left\lVert \hat{x} - x\right\rVert ^ 2 + s_x + e^{-s_q}\left\lVert \hat{q} - \frac{q}{\left\lVert q \right \rVert}\right\rVert ^ 2  + s_q.
\label{eq:geom_posenet_loss}
\end{equation}

This loss function was improper for the representation of directional statistics because quaternions lay on the unit sphere, and two opposite parts of this sphere corresponded to the same quaternion. For improving the loss function, Brahmbhatt et al. \cite{Brahmbhatt2018} proposed to use the logarithm of quaternion instead of quaternion in the loss function (equation \ref{eq:mapnet_loss}). Also, they changed the method for calculating the norm from $L_2$ to $L_1$.

\begin{equation}
\resizebox{0.9\columnwidth}{!}{$
\mathcal{L}(Y, \hat{X}) = e^{-s_x} \left\lVert \hat{x} - x\right\rVert _{1} + s_x + e^{-s_q}\left\lVert \log \hat{q} - \log  q \right\rVert _1  + s_q.
\label{eq:mapnet_loss}
$}
\end{equation}

These loss functions improved the results of the original PoseNet, but they still had problems. The first was the similarity of the $s_x$ and $s_q$ for each example from the training set. In other words, it was homoscedastic variance. Furthermore, the second one was the absence of the covariance between the rotation and position and between the different rotation and translation components.  

\subsection{The Loss Function based on the Lie Group}

We propose using the concentrated Gaussian distribution on the Lie group to model uncertainty in the loss function. This method is based on previous loss functions \ref{eq:mapnet_loss}, but this loss function has two distinctions. The first is the heteroscedastic variance assumption, leading to the heterogeneous loss function. We calculate the covariance matrix for the rotation and position as the output of neural network. The second improvement is changing the distribution from Gaussian to concentrated Gaussian distribution based on the Lie group. Thus, an output of the neural network is the mean and variance of the neural network (equations \ref{eq:mean_lie} and \ref{eq:variance_lie}): 

\begin{equation}
\mu = \mu_\phi(f_\theta(Y)),
\label{eq:mean_lie}
\end{equation}
\begin{equation}
\Sigma = \Sigma_\psi(f_\theta(Y)),
\label{eq:variance_lie}
\end{equation}

where $\mu$ is the mean of Lie distribution, $\Sigma$ is the variance of the Lie distribution, $f_\theta$ is image encoder, and $\phi$, $\theta$, $\psi$ are parameters of the neural network. The proposed loss function is represented by equation \ref{eq:lie_loss}. This equation is a negative logarithm of the concentrated Gaussian distribution from equations \ref{eq:mean_lie} and \ref{eq:variance_lie}:

\begin{equation}
\resizebox{0.9\columnwidth}{!}{$
\mathcal{L}(Y, X) = \frac{1}{2} \| \log_{SE(3)}(\mu^{-1} x) \|_\Sigma^2 + \frac{1}{2} \log |\Sigma| + N \log \sqrt{2 \pi}. 
$}
\label{eq:lie_loss}
\end{equation}

For numerical stability, we parametrize the variance of predicted distribution by Cholesky decomposition:

\begin{equation}
    \Sigma^{-1} = L L^T,
\end{equation}
where $L$ is the upper triangle matrix. This decomposition allows representing $L_2$ norm as $\| x \|_\Sigma^2 = \| L^{-1} x\|^2 $. This Cholesky factor is parametrized by 21 parameters: 6 parameters are for the diagonal, and 15 are for non-diagonal elements. 




The loss function of LiePoseNet has two advantages. Firstly, this loss function is invariant to rigid body transformations. Secondly, it assigns different weights for different images. In addition, The loss function of LiePoseNet can detect symmetry by increasing the weight in the covariance matrix. This way, it can assign a bigger weight to images that are most suitable for pose evaluation and do not have translational or rotational symmetry. This behavior allows us to speed up the learning process by skipping images with symmetry.

\section{Experimental Results}

\begin{table*}[]
\vspace{3ex}
\caption{Translation Error (m) and Rotation Error (\textdegree) for Different PoseNet-based Methods}
\label{table:main}
\resizebox{\textwidth}{!}{%
\begin{tabular}{|l|ccc|ccc|ccc|}
\hline
\rowcolor[HTML]{EFEFEF} 
Epochs &
  \multicolumn{3}{c|}{\cellcolor[HTML]{EFEFEF}300} &
  \multicolumn{3}{c|}{\cellcolor[HTML]{EFEFEF}100} &
  \multicolumn{3}{c|}{\cellcolor[HTML]{EFEFEF}10} \\ \hline
\rowcolor[HTML]{EFEFEF} 
Scene &
  \multicolumn{1}{c|}{\cellcolor[HTML]{EFEFEF}PoseNet17} &
  \multicolumn{1}{c|}{\cellcolor[HTML]{EFEFEF}PoseNet logq} &
  \multicolumn{1}{c|}{\cellcolor[HTML]{EFEFEF}LSTM PoseNet} &
  \multicolumn{1}{c|}{\cellcolor[HTML]{EFEFEF}PoseNet logq} &
  \multicolumn{1}{c|}{\cellcolor[HTML]{EFEFEF}\textbf{LiePoseNet L2}} &
  \multicolumn{1}{c|}{\cellcolor[HTML]{EFEFEF}\textbf{LiePoseNet L1}} &
  \multicolumn{1}{c|}{\cellcolor[HTML]{EFEFEF}PoseNet logq} &
  \multicolumn{1}{c|}{\cellcolor[HTML]{EFEFEF}\textbf{LiePoseNet L2}} &
  \multicolumn{1}{c|}{\cellcolor[HTML]{EFEFEF}\textbf{LiePoseNet L1}} \\ \hline
\cellcolor[HTML]{EFEFEF}Chess &
  0.13 / 4.48 &
  0.11 / 4.29 &
  0.24 / 5.77 &
  0.11 / 5.01 &
  0.17 / 5.60 &
  0.15 / 5.38 &
  0.19 / 9.11 &
  0.20 / 5.73 &
  0.18 / 5.39 \\ \hline
\cellcolor[HTML]{EFEFEF}Fire &
  0.27 / 11.30 &
  0.27 / 12.13 &
  0.34 / 11.9 &
  0.28 / 15.2 &
  0.35 / 11.24 &
  0.36 / 10.32 &
  0.30 / 19.58 &
  0.34 / 12.12 &
  0.36 / 10.05 \\ \hline
\cellcolor[HTML]{EFEFEF}Heads &
  0.17 / 13.00 &
  0.19 / 12.15 &
  0.21 / 13.7 &
  0.17 / 15.56 &
  0.24 / 15.13 &
  0.18 / 14.33 &
  0.19 / 22.92 &
  0.25 / 15.54 &
  0.20 / 14.86 \\ \hline
\cellcolor[HTML]{EFEFEF}Office &
  0.19 / 5.55 &
  0.19 / 6.35 &
  0.30 / 8.08 &
  0.18 / 6.99 &
  0.23 / 7.43 &
  0.23 / 6.78 &
  0.22 / 9.26 &
  0.25 / 7.34 &
  0.24 / 6.66 \\ \hline
\cellcolor[HTML]{EFEFEF}Pumpkin &
  0.26 / 4.75 &
  0.22 / 5.05 &
  0.33 / 7.00 &
  0.21 / 6.56 &
  0.30 / 7.01 &
  0.30 / 6.34 &
  0.24 / 10.0 &
  0.31 / 7.03 &
  0.29 / 6.06 \\ \hline
\cellcolor[HTML]{EFEFEF}Red Kitchen &
  0.23 / 5.35 &
  0.25 / 5.27 &
  0.37 / 8.83 &
  0.27 / 8.23 &
  0.29 / 7.07 &
  0.28 / 6.86 &
  0.29 / 9.94 &
  0.31 / 6.97 &
  0.30 / 6.66 \\ \hline
\cellcolor[HTML]{EFEFEF}Stairs &
  0.35 / 12.40 &
  0.30 / 11.29 &
  0.40 / 13.7 &
  0.39 / 15.63 &
  0.39 / 12.68 &
  0.36 / 11.98 &
  0.47 / 11.34 &
  0.36 / 13.57 &
  0.34 / 12.49 \\ \hline
\cellcolor[HTML]{EFEFEF}\textbf{Average} &
  0.23 / 8.12 &
  \textbf{0.22} / \textbf{8.07} &
  0.31 / 9.85 &
  \textbf{0.23} / 9.84 &
  0.28 / 9.45 &
  0.27 / \textbf{8.90} &
  0.27 / 14.59 &
  0.29 / 9.76 &
  \textbf{0.27} / \textbf{8.88} \\ \hline
\end{tabular}%
}
\vspace{-3ex}
\end{table*}

\subsection{Implementation Details}
Our implementation of PoseNet with Lie loss function (LiePoseNet) is based on the PoseNet (logq) implementation of Brahmbhatt et al. \cite{Brahmbhatt2018}\footnote{\url{https://github.com/NVlabs/geomapnet}}. For feature extraction, we use ResNet34 \cite{he2016} with pretrained weight on ImageNet, and for Lie geometry on PyTorch we use library with the open realization from GitHub repository\footnote{\url{https://github.com/utiasSTARS/liegroups}}. For optimization of the LiePoseNet model, we utilize Adam optimizer \cite{Kingma2015} with a learning rate of $1e-4$, weight decay of $5e-4$, and default betas. The optimization is performed during ten epochs. Input images are rescaled to $256 \times 341$. The initial values for rotation and translation coefficients are set $s = -3$.

We find that the implementation of PoseNet (logq) has an error with dropout usage. It has a zero dropout rate for all models. We have fixed this problem and conduct an ablation study to determine the best dropout rate of the model with the heterogeneous Lie loss function. The models with zero dropout rates performs better than those with non-zero dropout. Therefore, for the comparative study, we use a zero dropout rate.

\subsection{Dataset}
For evaluation of our method, we use the 7-Scenes dataset \cite{shotton2013}. This dataset consists of color and depth images from different scenes inside a building. Each scene is further divided into several sequences, with 1000 images in each sequence. The ground truth trajectories are obtained from KinectFusion by a SLAM algorithm for all sequences. In addition, all sequences are divided into two categories. Sequences from the first category are used for training and from the second for the model evaluation.

\subsection{Comparison with the Previous Works}
For the evaluation of our approach, we use rotation and translation errors. Median is commonly used for the aggregation of these metrics. These metrics are calculated for each scene and then averaged for all scenes for comparison. For a comparative experiment (Table \ref{table:main}), we choose the following works: PoseNet17 \cite{Kendall2017}, Spatial LSTM PoseNet \cite{walch2017} and PoseNet (logq) with loss function based on logarithm of quaternion \cite{Brahmbhatt2018}.


Firstly, we calculate the accuracy of the PoseNet (logq) method \cite{Brahmbhatt2018} with training during 10 epochs to understand whether this method is enough for training with an acceptable error in a short time. This method's average accuracy is 14.59 degrees of median rotation error and 0.27 meters of median translation error. This result is worse than the result of the spatial LSTM PoseNet method. It means that more than 10 epochs are needed to train the PoseNet (logq) method \cite{Brahmbhatt2018}, and it could not speed up the training process significantly. Therefore, it is necessary to use another method to speed up the training process, for example, presented in this work.

\begin{figure} [!t]
\vspace{2ex}
\begin{center}
\includegraphics[width=8.6 cm]{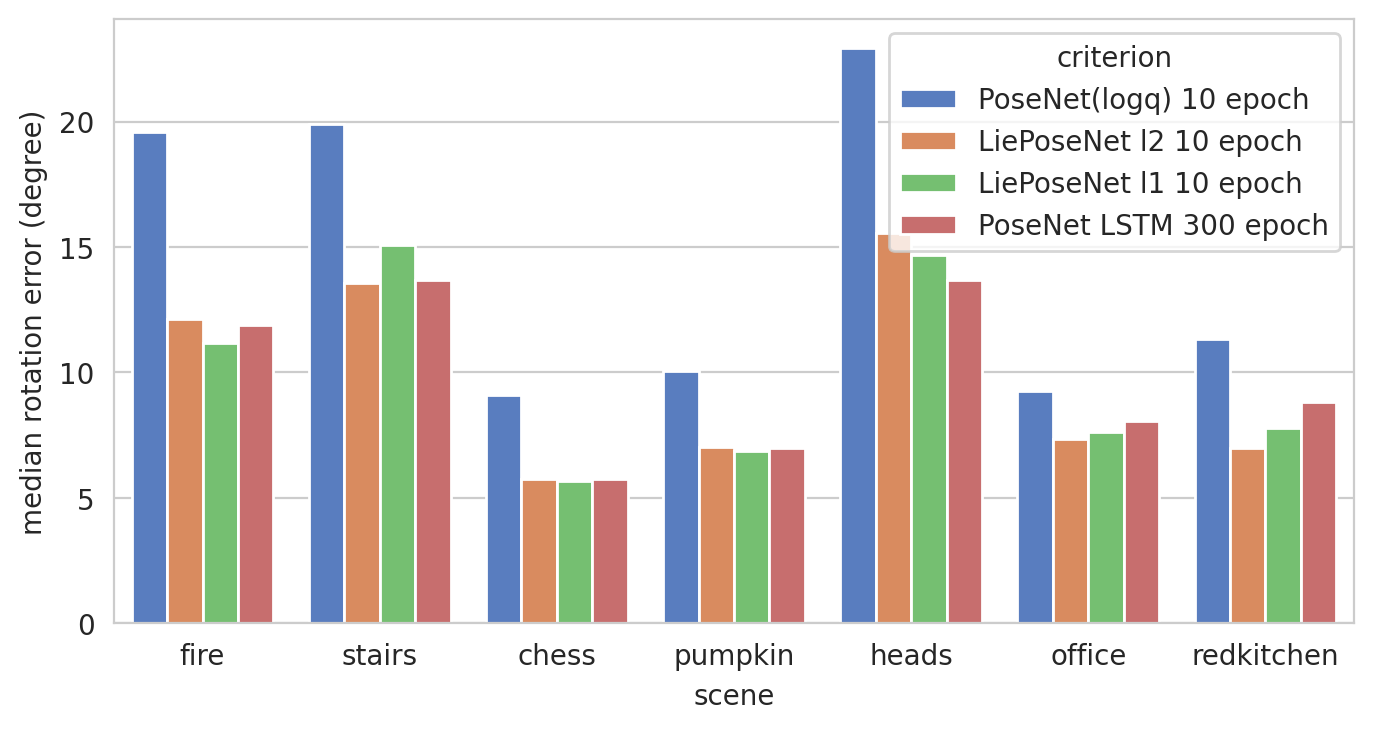}
\vspace{-5ex}

\caption{Median rotation error (\textdegree) for ten epochs.}
\label{fig:median_rotation_10epoch_bar}
\end{center}
\vspace{-6ex}
\end{figure}

Moreover, a bar chart for 10 epochs is presented in Fig. \ref{fig:median_rotation_10epoch_bar}. According to the obtained results, LiePoseNet has a significantly better median rotation error than PoseNet (logq). The average median rotation error for LiePoseNet with L2 norm is 9.76\textdegree \, and for LiePoseNet with L1 norm, the median rotation error is 8.88\textdegree. The average median rotation error for the whole training process during 100 epochs is presented in Fig. \ref{fig:median_rotation_error_plot}. Following Fig. \ref{fig:median_rotation_error_plot}, the median rotation error for LiePoseNet is smaller than for PoseNet.

Moreover, LiePoseNet has better performance than the spatial LSTM PoseNet method. The obtained result highlights that our framework significantly reduces (from 300 to 10 epochs) training time with accuracy no less than the spatial LSTM PoseNet method. During experiments, we observe that training of 300 epochs for PoseNet17 takes 107 minutes. In contrast, our training of 10 epochs for our method takes 4 minutes on the same hardware setup that 26.75 times faster.

\begin{figure} [!t]
\begin{center}
\includegraphics[width=8.6 cm]{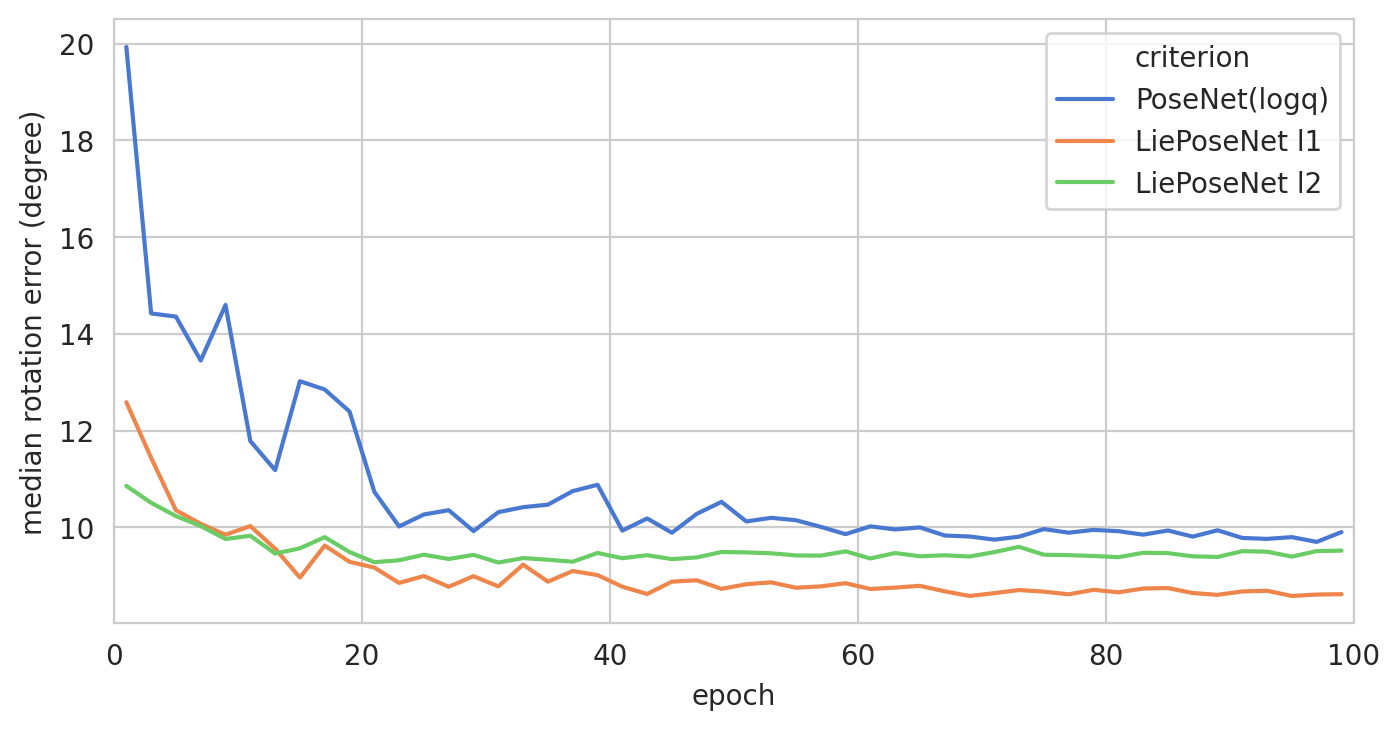}
\vspace{-5ex}

\caption{Median rotation error (\textdegree) averaged for all 7-Scenes.}
\label{fig:median_rotation_error_plot}
\end{center}
\vspace{-6ex}
\end{figure}
\section{Conclusion and Discussion}

In this work, we propose the LiePoseNet loss function approach. This method uses the loss function based on heteroscedastic variance assumption and concentrated Gaussian distribution based on the Lie group. 

We compare the result of the LiePoseNet trained during 10 epochs and PoseNet (logq) with the logarithm quaternion loss function. We receive that the LiePoseNet has the best result for rotation and translation errors after ten epochs of training. We also find out that our loss function allows speeding up the training process 30 times (from 300 to 10 epochs) with acceptable error values and even receives the best result of overall rotation error after training in comparison with all implemented methods.

During experiments, we show that the effect of the LiePoseNet loss function no longer influences after 20 epochs of training, where it reaches minimum error. Thus, this method allows getting the result much faster and test hypotheses earlier, and also affects on quick adaption to new environment changes. However, the maximum accuracy could not be increased by this method.

Nonetheless, our result is not accurate enough compared to PoseNet trained during 300 epochs. Our further research will be devoted to improving the result of LiePoseNet localization. We will use a projective geometry loss function and its combination with Lie group.

\section{Acknowledgements}
The reported study was funded by RFBR and CNRS according to the research project No. 21-58-15006.











\bibliographystyle{IEEEtran}
\bibliography{literature.bib}

\end{document}